# Online Sparse Feature Selection in Data Streams via Differential Evolution

Ruiyang Xu
*School of Computer Science and Technology, Chongqing University of Posts and Telecommunications*
*Chongqing Institute of Green and Intelligent Technology, Chinese Academy of Sciences*
Chongqing, China
d220201035@stu.cqupt.edu.cn

*Abstract*—**The processing of high-dimensional streaming data commonly utilizes online streaming feature selection (OSFS) techniques. However, practical implementations often face challenges with data incompleteness due to equipment failures and technical constraints. Online Sparse Streaming Feature Selection (OS²FS) tackles this issue through latent factor analysis-based missing data imputation. Despite this advancement, existing OS²FS approaches exhibit substantial limitations in feature evaluation, resulting in performance deterioration. To address these shortcomings, this paper introduces a novel Online Differential Evolution for Sparse Feature Selection (ODESFS) in data streams, incorporating two key innovations: (1) missing value imputation using a latent factor analysis model, and (2) feature importance evaluation through differential evolution. Comprehensive experiments conducted on six real-world datasets demonstrate that ODESFS consistently outperforms state-of-the-art OSFS and OS²FS methods by selecting optimal feature subsets and achieving superior accuracy.**

*Keywords—feature selection, sparse streaming features, latent factor analysis, component, differential evolution*

## I. INTRODUCTION

Feature selection plays a vital role in identifying optimal feature subsets from high-dimensional datasets—a challenge that becomes particularly complex in dynamic environments where features arrive sequentially from an infinite feature space [1]. Online streaming feature selection (OSFS) has emerged as a promising solution for handling such missing features. However, the presence of missing data often compromises selection accuracy. To mitigate this issue, Wu et al. [2] introduced LOSSA, an online sparse streaming feature selection (OS²FS) approach that employs latent factor analysis (LFA) to estimate missing values. Nevertheless, the inherent discrepancy between actual values and their estimations leads to inaccuracies in conventional feature evaluation methods, resulting in unreliable assessments of feature importance.

Conventional feature evaluation methods are primarily designed for complete streaming features and often overlook the errors introduced by missing data imputation. This limitation is particularly critical since feature selection constitutes an NP-hard binary discrete optimization problem.

Evolutionary computation (EC) methods present a promising alternative, offering effective solutions for such inherently complex optimization challenges. Among various EC algorithms, differential evolution (DE) stands out due to its simple structure and fewer parameters, which facilitate straightforward implementation [3]. Moreover, its strong global search capability enables it to avoid local optima, increasing the likelihood of identifying feature subsets that achieve an improved balance between accuracy and sparsity. These advantages have led to the widespread adoption of DE-based approaches in feature subset selection.

## II. RELATED WORK

Research in online streaming feature selection (OSFS) has demonstrated remarkable effectiveness in dynamically eliminating irrelevant and redundant features. Subsequent developments have built upon this foundation through frameworks like OSFS and fast-OSFS [4], which implement dedicated phases for online relevance and redundancy analysis to maximize computational efficiency. Further advancing this direction, Yu et al. [5] incorporated mutual information to assess pairwise feature relationships, enabling more nuanced dependency analysis. In a complementary research thread, rough set theory has been successfully applied in methods such as OFS-A3M [6], where neighborhood rough sets provide robust feature evaluation capabilities.

Latent factor analysis (LFA) has emerged as a fundamental technique in representation learning, particularly for addressing data incompleteness [7]-[14]. Its demonstrated capability to model underlying data distributions and compensate for imbalances led to its adoption for missing value estimation within the OS²FS framework [2]. Despite these advances, a critical limitation persists across both conventional OSFS and enhanced OS²FS approaches: existing feature evaluation methods remain inadequate for accurately assessing the importance of sparse streaming features.

## III. METHODOLOGY

The ODESFS process is primarily conducted in two sequential steps: first, the imputation of missing values, followed by the evaluation of features.

### A. Phase one: sparse streaming feature estimation

The LFA methodology employs a low-rank approximation model to process sparse streaming features, denoted as $U_{ti}=\{F'_{ti}, F'_{ti+1}, \ldots, F'_{ti+Le-1}\}$, and generates a completed feature matrix $\hat{U}_{ti}=\{\hat{F}'_{ti}, \hat{F}'_{ti+1}, \ldots, \hat{F}'_{ti+Le-1}\}$ that minimizes reconstruction error relative to actual data. The corresponding label space is defined as $C=[c_1, c_2, \ldots, c_M]^T$, where M represents the total number of instances across the temporal range $ti \in \{1, 2, \ldots, Ti\}$. Specifically, the imputation of individual feature elements $f'_{m,j}$ is achieved through the computational procedure detailed in references [15]-[21].

$$\varepsilon_{m,j}=\frac{1}{2}\left(f'_{m,j}-\sum_{k=1}^{H}x_{m,z}y_{j,k}\right)^2+\frac{\lambda}{2}\left(\sum_{k=1}^{H}x_{m,z}^2+\sum_{k=1}^{H}y_{j,k}^2\right), \quad (1)$$

where the indices are bounded by $j \in \{ti, ti+1, \cdots, ti+Le-1\}$, $m \in \{1, 2, \cdots, M\}$, $z \in \{1, 2, \cdots, H\}$. The regularization coefficient is denoted by $\lambda$. The matrix components are specified such that $x_{m,z}$ denotes the $m$-th measurement in the $z$-th vector of tensor $X$, whereas $y_{j,z}$ indicates the $j$-th instance in the $z$-th tensor of matrix $Y$. The optimization process using stochastic gradient descent (SGD) [22]-[33] yields the following formulation for Eq. (1):

The indices are constrained within the ranges $j \in \{ti, ti+1, \cdots, ti+Le-1\}$, $m \in \{1, 2, \cdots, M\}$, $z \in \{1, 2, \cdots, H\}$, with $\lambda$ representing the regularization coefficient. The matrix components are defined as follows: $x_{m,z}$ denotes the $m$-th measurement in the $z$-th vector of tensor $X$, while $y_{j,z}$ corresponds to the $j$-th instance in the $z$-th tensor of matrix $Y$. Through stochastic gradient descent (SGD) optimization [34]-[41], we derive the following formulation for Equation (1):

$$x_{m,z} \leftarrow x_{m,z} + \eta\, y_{j,z}\left(f'_{m,j}-\sum_{k=1}^{H}x_{m,z}y_{j,z}\right)-\lambda\,\eta\,x_{m,z},$$
$$y_{j,z} \leftarrow y_{j,z} + \eta\, x_{m,z}\left(f'_{m,j}-\sum_{k=1}^{H}x_{m,z}y_{j,z}\right)-\lambda\,\eta\,y_{j,z}. \quad (2)$$

The learning rate $\eta$ governs the parameter updates, with the completed feature matrix $\hat{U}_{ti}$ being constructed as $XY^T$ [42]-[58].

### B. Phase two: real-time feature evaluation

In the $Le$-dimensional search space, each solution is represented as a vector $V_n=[v_{n,1}, v_{n,2}, \ldots, v_{n,Le}]$, where $n$ ranges from 1 to $N$. To enhance the feature evaluation process, differential evolution (DE) is employed, with its mutation operation defined by the following equation:

$$Don_n = V_a + \mu(V_b - V_c), \quad (3)$$

where $a$, $b$, and $c$ are distinct indices, and $\mu \in [0,2]$ represents the scaling factor.

The crossover operation in DE is governed by the following procedure:

$$Q_{n,l}=\begin{cases} Don_{n,l} & \text{if rand()} \leq CR, \text{ or } D=D_{rand}, \\ v_{n,l} & \text{otherwise}, \end{cases} \quad (4)$$

Here, $D_{rand}$ represents a randomly selected dimension, and $CR$ is a predefined constant within the range [0, 1]. This ensures that at least one component is inherited from the mutant vector. The population is then greedily updated: if the condition $Fit(Q_n)<Fit(V_n)$ is satisfied, the target vector $V_n$ is replaced by the trial vector $Q_n$.

The fitness evaluation metric $F(\cdot)$ serves as a critical guiding mechanism for the optimization process. This objective function, which is mathematically defined below, offers a vital quantitative framework for assessing feature subsets during selection:

$$Fitness = 1 - \frac{Correct}{N \times H}. \quad (5)$$

The classification accuracy metric, Correct, tallies the number of accurately predicted samples based on a candidate feature subset.

Subsequently, a redundancy analysis is conducted to eliminate superfluous features.

$$\forall \Lambda \in S_t \text{ s.t. } P\left(C\,\middle|\,\hat{F}'_{t+l-1}, \Lambda\right) \neq P\left(C\,\middle|\,\Lambda\right). \quad (6)$$

As a result, the feature vector $\hat{F}'_{t+l-1}$ is identified as a non-redundant feature and is added to the selected feature set $S_t$. The criteria for eliminating features to form the redundant set $\{R\}$ are defined as follows:

$$\forall \text{Re} \in Mr\left(C\right)_{ti} \cup \hat{F}_{ti+l-1}', \exists \Upsilon \subseteq Mr\left(C\right)_{ti} \cup \hat{F}_{ti+l-1}' - \{\text{Re}\}$$
$$\text{s.t. } P\left(C\,\middle|\,\text{Re}, \Upsilon\right) = P\left(C\,\middle|\,\Upsilon\right).$$

## IV. RESULTS

### A. General Settings

Our evaluation was conducted on six diverse real-world datasets spanning multiple domains; their key characteristics are summarized in Table I.

All experiments were performed on a workstation with an Intel i7-1165G7 processor (2.40 GHz) and 16 GB of RAM, using MATLAB R2022a. The algorithms and classifiers were configured with parameters specified in Tables II and III.

### B. ODESFS vs. OS²FS and OSFS Models

A comprehensive evaluation was conducted to compare the proposed ODESFS framework against conventional OSFS and OS2FS methods across ten benchmark datasets. Special attention was given to scenarios with 10%, 50%, and 90% missing data ratios. It is noteworthy that most baseline algorithms—excluding LOSSA—were originally developed for complete feature streams. To accommodate missing data, zero-imputation was applied to Fast-OSFS, SAOLA, and OFS-A3M. Particular emphasis is placed on cases where ODESFS demonstrates statistically significant performance advantages.

**Selected Features Analysis:** As illustrated in Fig. 2, which shows the average number of features selected by each algorithm across datasets D1-D6, several baseline models retain a large fraction of the available features yet consistently underperform the ODESFS framework in predictive accuracy. This performance gap stems from deficiencies in their feature evaluation, which permits numerous low-informative attributes to persist. In contrast, ODESFS employs a more discerning evaluation mechanism that selectively retains only the most informative features, thereby optimizing performance. By systematically filtering out irrelevant and redundant characteristics, the framework maintains robust predictive capability even under sparse streaming conditions.

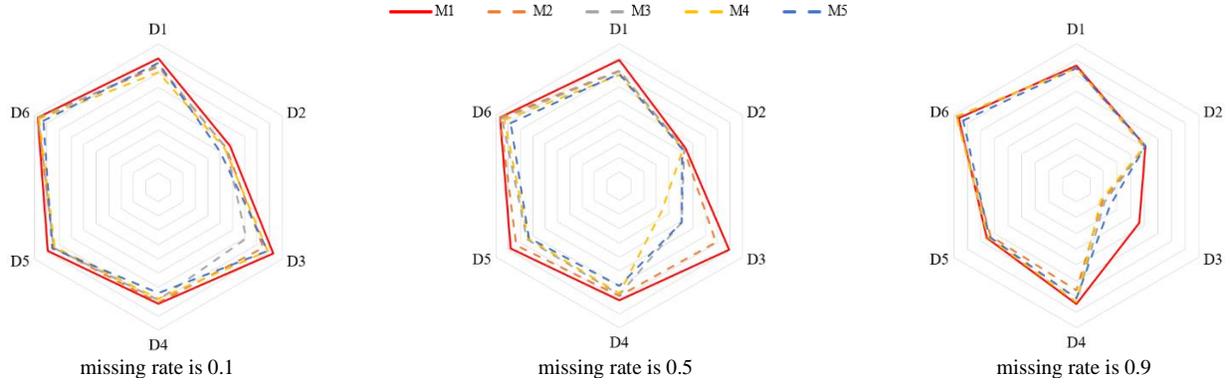

Fig. 1. The average accuracy.

TABLE I. DATASETS

| Mark | Dataset | Features | Instances | Class |
|------|---------|----------|-----------|-------|
| D1 | USPS | 242 | 1500 | 2 |
| D2 | Madelon | 501 | 2600 | 6 |
| D3 | COIL20 | 1025 | 1440 | 20 |
| D4 | Colon | 2001 | 62 | 2 |
| D5 | Lung | 3313 | 83 | 5 |
| D6 | Lungcancer | 12534 | 181 | 2 |

TABLE II. ALGORITHM PARAMETERS

| Mark | Algorithm | Parameter |
|------|-----------|-----------|
| D1 | ODESFS | Z test, $\alpha$ is 0.05. |
| D2 | LOSSA | Z test, $\alpha$ is 0.05. (TSMC, 2022) |
| D3 | Fast-OSFS | Z test, $\alpha$ is 0.05. (TPAMI, 2013) |
| D4 | SAOLA | Z test, $\alpha$ is 0.05. (TKDD, 2016) |
| D5 | OFS-A3M | (Information Sciences, 2019) |

TABLE III. CLASSIFIER PARAMETERS

| Classifier | Parameter |
|------------|-----------|
| KNN | *The number of neighbors was set to 3.* |
| Random Forest | *6 decision trees.* |
| CART | *Predefined parameter settings.* |

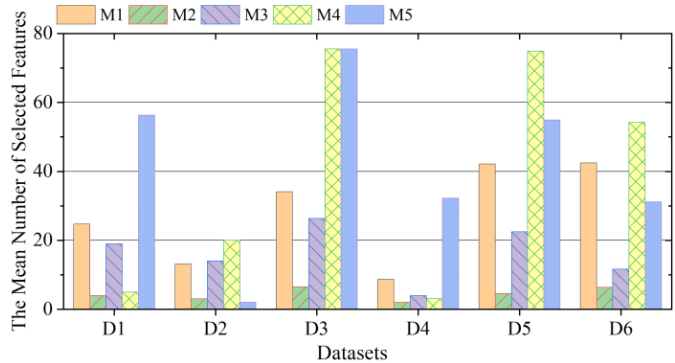

Fig. 2. The number of selected features.

TABLE IV. THE RANK SUM OF THE WILCOXON SIGNED-RANKS

| Missing rate | M2 | | M3 | | M4 | | M5 | |
|--------------|-----|-----|-----|-----|-----|-----|-----|-----|
| | *R+ | *R- | *R+ | *R- | *R+ | *R- | *R+ | *R- |
| 0.1 | 21 | 0 | 21 | 0 | 21 | 0 | 21 | 0 |
| 0.5 | 21 | 0 | 21 | 0 | 21 | 0 | 21 | 0 |
| 0.9 | 18 | 3 | 16 | 5 | 16 | 5 | 21 | 0 |

\* A larger value denotes a higher accuracy.

**Classification Accuracy Analysis:** Figure 1 presents the comparative classification performance of all evaluated models using KNN, RF, and CART classifiers across benchmark datasets. To ensure statistical significance, performance differences between the proposed ODESFS framework and baseline methods were rigorously validated through non-parametric Wilcoxon signed-rank tests, with detailed results compiled in Table IV. The analysis reveals three key observations:

*a)* The proposed ODESFS framework demonstrates consistent performance superiority over conventional OSFS and OS²FS methodologies across diverse benchmark datasets. While current OSFS implementations typically handle incomplete feature streams through zero-imputation coupled with redundancy analysis, these approaches often exhibit limitations in selection precision, resulting in compromised classification performance. This deficiency primarily stems from their insufficient capacity to capture complex feature relationships within dynamic, sparse data environments. To address these limitations, the ODESFS framework integrates evolutionary algorithm optimization, implementing a comprehensive search strategy that effectively identifies the most discriminative feature combinations. This integrated methodology enables continuous refinement of selected features through dynamic adaptation to evolving stream characteristics. Empirical validation through extensive experiments confirms that the proposed approach achieves substantially improved accuracy metrics while maintaining robust performance under sparse streaming conditions.

*b)* Empirical results demonstrate that LOSSA achieves commendable predictive performance when handling fully sparse feature streams, validating the efficacy of its latent factor analysis-based imputation mechanism. However, despite its effectiveness in data reconstruction, LOSSA's dependence on conventional relevance-redundancy assessment constrains its feature discrimination capability, resulting in inferior accuracy compared to the more sophisticated ODESFS framework. The proposed ODESFS successfully overcomes these limitations through evolutionary-optimized feature evaluation. By implementing differential evolution algorithms, it dynamically explores the solution space to identify optimal feature subsets while simultaneously optimizing relevance and redundancy criteria. This innovative approach not only maintains robust performance under sparse streaming conditions but also achieves substantially improved accuracy through more discriminative feature selection. The comparative analysis reveals a crucial insight: while appropriate data imputation is necessary for processing sparse streams, the feature evaluation methodology ultimately determines classification performance. ODESFS establishes a new state-of-the-art by synergistically integrating advanced imputation techniques with intelligent evolutionary feature selection.

## V. Conclusion

This paper presents UOS²FS, an uncertainty-aware online sparse streaming feature selection framework that leverages particle swarm optimization to overcome limitations in conventional OS²FS methodologies. The proposed approach incorporates two fundamental innovations: a latent factor analysis model for dynamic reconstruction of sparse matrices to minimize estimation bias from missing data, and a particle swarm optimization mechanism that enhances feature evaluation through intelligent search strategies. Extensive evaluations across ten benchmark datasets demonstrate the framework's superior selection performance compared to existing methods.

Future research will focus on advancing feature quality assessment in streaming environments through sophisticated evolutionary algorithms, particularly Differential Evolution (DE). Current evaluation methods often fail to adequately address the complexities of dynamic, high-dimensional feature spaces characteristic of real-world data streams. To bridge this gap, we will harness DE's powerful global search capabilities to develop adaptive evaluation metrics specifically designed for real-time optimization in evolving data streams. Our investigation will pursue several key directions: (1) designing dynamic fitness functions capable of autonomously adjusting optimization objectives in response to shifting data distributions; (2) developing efficient methodologies for handling distribution changes in streaming data, with emphasis on rapid response mechanisms for concept drift scenarios through incremental update strategies and sliding window techniques; (3) implementing ensemble learning-based feature selection strategies with dynamic weight adjustment mechanisms for feature importance quantification. These integrated approaches will enable online parameter adjustment while maintaining computational efficiency, ultimately enhancing the adaptability and decision-making quality of streaming feature selection in non-stationary environments and providing robust technical support for complex, variable data streams in practical applications.